\documentclass{article}
\usepackage{graphicx} 
\usepackage[a4paper,margin=1in]{geometry}
\usepackage[
backend=biber,
style=ieee,
sorting=none,
giveninits=false,
maxnames=99  
]{biblatex}
\usepackage[hidelinks]{hyperref} 
\title{Targeted Manipulation: Slope-Based Attacks on Financial Time-eries Data}
\author{Dominik Luszczynski}
\date{June 2025}
\usepackage{graphicx}
\usepackage{amsmath}
\usepackage{amsthm}
\usepackage{amssymb}
\usepackage{indentfirst}
\usepackage{enumerate}
\usepackage{subcaption}
\usepackage{graphicx}
\usepackage{tikz}
\usepackage{multirow}
\usepackage{makecell}
\usepackage{algorithm}
\usepackage{algpseudocode}
\usepackage{algorithmicx}
\usepackage{float}
\usepackage[labelformat=simple]{caption}
\usepackage{ragged2e} 
\usepackage{placeins}
\usepackage{fontawesome5}
\justifying
\begin{document}
\noindent \rule{\textwidth}{2pt}

\begin{center}
    \noindent \textbf{\LARGE Targeted Manipulation: Slope-Based Attacks on Financial Time-Series Data}
    \noindent \rule{\textwidth}{0.5pt}
\end{center}

\begin{center}
    \textbf{Dominik Luszczynski$^{1}$}
\end{center}

\noindent $^{1}$ University of Toronto Department of Computer Science \hfill Toronto, ON, Canada\\
\begin{center}
\textsuperscript{\faEnvelope[regular]} \texttt{dominik.luszczynski@mail.utoronto.ca}\\
\end{center}


\section*{Abstract}
A common method of attacking deep learning models is through adversarial attacks, which occur when an attacker specifically modifies the input of a model to produce an incorrect result. Adversarial attacks have been deeply investigated in the image domain; however, there is less research in the time-series domain and very little for forecasting financial data. To address these concerns, this study aims to build upon previous research on adversarial attacks for time-series data by introducing two new slope-based methods aimed to alter the trends of the predicted stock forecast generated by an N-HiTS model. Compared to the normal N-HiTS predictions, the two new slope-based methods, the General Slope Attack and Least-Squares Slope Attack, can manipulate N-HiTS predictions by doubling the slope. These new slope attacks can bypass standard security mechanisms, such as a discriminator that filters real and perturbed inputs, reducing a 4-layered CNN's specificity to 28\% and accuracy to 57\%. Furthermore, the slope based methods were incorporated into a GAN architecture as a means of generating realistic synthetic data, while simultaneously fooling the model. Finally, this paper also proposes a sample malware designed to inject an adversarial attack in the model inference library, proving that ML-security research should not only focus on making the model safe, but also securing the entire pipeline.

\section*{Key Words}
Adversarial Attacks, GANs, ML Security, N-HiTS

\section{Introduction}
The increased use of machine learning (ML) and artificial intelligence (AI) models in high-risk sectors, such as healthcare and finance, make it imperative that these predictive models are robust. It is known that deep neural networks (DNN) are highly susceptible to adversarial attacks \cite{mifgsm:2018}. Adversarial attacks occur when an attacker slightly modifies the input to a model, typically by adding noise, and causes it to produce an incorrect result \cite{fgsm:2014}. Adversarial attacks are extremely problematic in noise sensitive domains, such as healthcare, finance and computer vision. For example, a reliable and safe self-driving car must be able to identify a stop sign, even if some noise has been added to its vision. Adversarial attacks come in two forms: white and black-box attacks. A white box attack occurs when the attacker has full access to all information about the model, enabling the attacker to exploit gradient information \cite{tar_and_stealthy:2025}. A black-box attack occurs when the model is hidden from the attacker, and they do not have any knowledge about the structure or parameters, and do not have access to gradient information \cite{healthcare:2022, tar_and_stealthy:2025}.\\

Most contributions to adversarial attack research has been focused on image and text classification, while research about attacks on time-series data, especially financial data, is still in its early stages \cite{tar_and_stealthy:2025}. There have been recent advancements in the time-series domain, with Gallager et al., attacking a simple three layered Convolutional Neural Network (CNN) forecasting the Google Stock from 2006 to 2018 with Fast Gradient Sign Method \cite{gallager_et_al:2022}. Moreover, Rathore et al., applied the Fast Gradient Sign Method and the Basic Iterative Method on a classification model trained on 54 different time-series datasets related to healthcare, vehicle sensors and electrical equipment \cite{rathore:2018}. Rathore et al., also discovered that time-series classification models are susceptible to adversarial attacks and defense strategies, such as adversarial training (including adversarial examples in the training set), are effective \cite{rathore:2018}. The introductory research conducted shows how susceptible time-series models are to adversarial attacks; however, they typically focus on classification problems and small-scale models, and do not consider the believability of an adversarial example \cite{time_series_adv_review:2025, tar_and_stealthy:2025}.\\

A significant challenge regarding adversarial attacks in the time-series domain is that, unlike images, perturbations made on time-series are more noticeable. To combat this challenge, Shen and Li introduced Stealthy attacks, which rely on using cosine similarities, to ensure temporal characteristics of time-series remain intact \cite{tar_and_stealthy:2025}. Furthermore, Pialla et al., introduced the Smooth Gradient Method which involves constraining the noise optimization by a regularization term penalizing non-smoothness \cite{pialla:2025}. As a result, these attack methods are able to better bypass human and machine detection; however, there is a tradeoff between the amount of error caused by the attack versus making it look believable. \\

In an attempt to generate more believable attack inputs, Generative Adversarial Networks (GANs) have been used to create adversarial attacks \cite{maggan:2020}. GANs typically comprise of two networks, a generator and a discriminator (or critic), where the generator creates synthetic data, while the discriminator determines whether it is real or fake \cite{timeGanImprovement:2025}. At the end of training, the GAN should be able to generate believable synthetic data that would ideally bypass detection. Researchers have experimented with adding a second critic to the GAN architecture, which is the model that the attacker is aiming to fool \cite{maggan:2020}. However, similar to adversarial attacks, the majority of research on GANs have been on images, and there have been even fewer studies using GANs to generate adversarial examples for time-series forecasting \cite{maggan:2020, psat_gan:2021, adv_gan:2024}.\\

Furthermore, many adversarial attacks aim to maximize the error produced by the model; however, this is insufficient for an attacker since it does not lead to any clear advantage. Instead, targeted adversarial attacks aim to force the model to predict a specific output which the attacker desires, improving the applicability of the attack. Previous targeted adversarial attacks, such as the attacks developed by Shen and Li, aim to minimize the loss between the target and the prediction, where the target is either the real data modified by some margin or a new sequence $Y^*$ \cite{tar_and_stealthy:2025}. While this is an effective strategy, it is unrealistic for an attacker to generate a new sequence $Y^*$ for different inputs and, modifying with a scalar margin would only shift the predictions, leading to poor applicability for an attacker as the temporal characteristics are left unchanged. The limitations of current targeted methods is further exacerbated in the financial time-series domain, and given that stock data is extremely volatile and noisy, it is necessary for the target to be embedded inside the objective function. \\

\begin{figure}[H]
    \centering
    \includegraphics[scale=0.5]{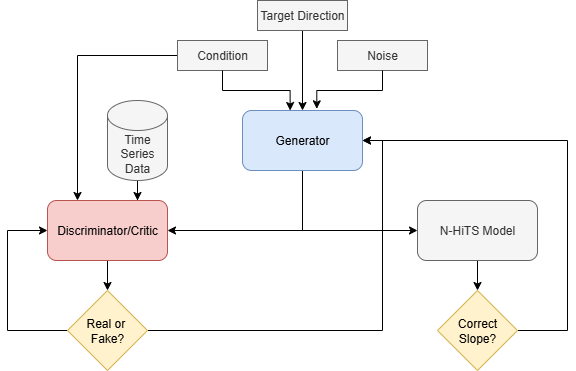}
    \vspace{-2mm}
    \caption{High-level architecture diagram of the proposed targeted adversarial GAN}
    \label{fig:adv_gan_architecture}
\end{figure}

This research aims to build upon previous studies on adversarial attacks by introducing two slope-based targeted attacks on financial time-series data, aimed to alter the temporal characteristics of a model's predictions. Furthermore, this paper examines the ability of GANs to generate adversarial examples, based on a slope-based objective. The use of GANs enables an attacker to generate realistic and believable time-series data, ideally without sacrificing how much error the model produces. Using an adversarial GAN also allows the mass generation of synthetic data for adversarial training, which is a proven method to help improve model robustness \cite{pialla:2025}. Moreover, existing adversarial attacks made for time-series data are performed on relatively simple architectures, such as a shallow CNN \cite{tar_and_stealthy:2025, gallager_et_al:2022}. Thus, this study attempts to target more state-of-the-art forecasting models, such as the N-HiTS model \cite{nhits:2022}. However, rather than training a model to forecast a single stock, the N-HiTS model is trained on over 300 recordings of stock data to improve the robustness of the model, thus making adversarial attacks more difficult. Figure \ref{fig:adv_gan_architecture} provides a visual display of the adversarial GAN architecture. Lastly, this research aims to investigate existing adversarial attack strategies and their performance on the N-HiTS architecture. 

\section{Forecasting Model}
\subsection{Dataset and Preprocessing}
All financial stock data was collected from the Center for Research Security Prices (CRSP). Specifically, the daily adjusted price (adjprc) of all S\&P 500 stocks was collected from CRSP for the 2015:05:01-2025:05:01 period. To determine the training, validation and testing set split, a stratified split was made by creating eight evenly distributed bins based on the "Stock Price" collected from \cite{stockAnalysis:2025}. The training, validation and testing set comprises of 360, 48, and 72 recordings respectively, or equivalently 75\%, 10\% and 15\% of the dataset. Any stocks with less than 600 days of recordings were omitted. In order to enrich the feature set for the forecasting model, several features were derived from the adjprc:
\begin{itemize}
    \item Rolling mean with window sizes of 5, 10, 20
    \item Rolling standard deviation with window sizes of 5, 10, 20
    \item Log Returns
    \item Rate of Change with a delta of 5
    \item Exponential moving averages with window sizes 5, 10 and 20 
\end{itemize}
Furthermore, from the date of each adjprc, the day of the week was extracted as a categorical feature because it has been shown that there are differences in volatility between the beginning and end of the trading week \cite{day_of_week_effect:2017}.  In order to preserve the computational graph for adversarial attacks, all feature generation was performed in PyTorch.
\subsection{N-HiTS Model}
N-HiTS is a novel projection model which builds upon the N-BEATS architecture, simultaneously improving computational performance and accuracy by sampling the time-series at different rates \cite{nhits:2022}. The N-HiTS architecture uses Multi-layer Perceptrons (MLPs) for each block of the time-series to estimate coefficients for the backcasts and forecasts \cite{nhits:2022}. The backcasts are used to clean future blocks, while the forecasts are summed to generate the predictions \cite{nhits:2022}. However, the N-HiTS model introduces novel components to each block, like using MaxPool with some kernel size $k_b$ for each block $b$, claiming that it helps the MLP focus on low frequency and large scale contents of the time-series, which allows it to sample different rates \cite{nhits:2022}. After the pooling layer, a non-linear regression is applied to estimate the backcasts and forecasts, and to eventually generate the final predictions, hierarchal interpolation is performed \cite{nhits:2022}. Given that N-HiTS is making predictions at different frequencies, the hierarchal interpolation assigns each stack of predictions with an expressiveness ratio, representing the number of predictions in the stack, then sums the stacks of predictions according to the magnitude of the expressiveness ratio \cite{nhits:2022}.\\

The N-HiTS model was implemented through the PyTorch-Forecasting library. The model was trained for 100 epochs, with an encoder length of 100 and a prediction length of 20, meaning it uses the previous 100 days of data to forecast the next 20. The hyperparameters used for the N-HiTS model are stated in Table \ref{tab:N-HiTS_hparams} in the Appendix. A PyTorch Lightning Tuner was used to determine the optimal learning rate, while weight decay and the hidden size were determined experimentally. Batch normalization and early stopping with a patience of 15 were used to help prevent overfitting. Several loss functions were experimented with, such as Mean Squared Error (MSE), L1 Loss, Mean Absolute Percentage Error (MAPE), and Symmetric Mean Absolute Percentage Error (SMAPE); however, Quantile Loss with quantiles 0.01, 0.1, 0.05, 0.5, 0.95, 0.99, and 0.999 gave the best results.

\subsection{N-HiTS Results}
\begin{figure}[H]
    \centering
    \includegraphics[scale=0.35]{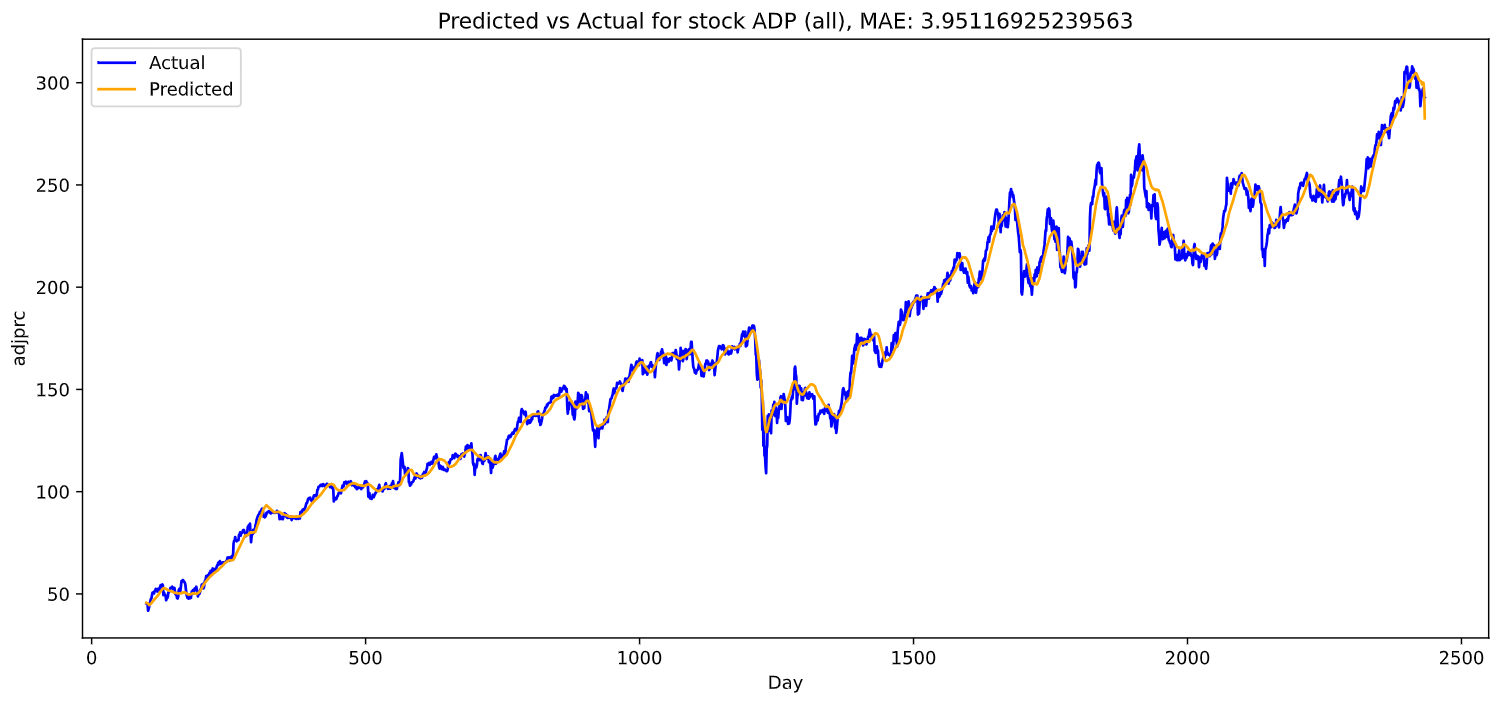}
    \caption{Example forecast for the \textit{ADP} Stock generated by the N-HiTS model}
    \label{fig:dhi_stock_projection}
\end{figure}

Using the test set, results were computed by comparing the true adjprc with the adjprc forecast generated from the N-HiTS model. The predictions were made on the entire recording starting from the 100th day, as the N-HiTS model requires a look-back period of 100 days. Given that the model produces 20 day forecasts, the N-HiTS model makes predictions over a rolling window of 100 days. Corresponding predictions on the same day are then averaged for the final prediction. The model achieved an average Mean Absolute Error (MAE) of 4.11, a Root Mean Squared Error of 6.04 and a Mean Absolute Percentage Error of 3.48\%. To illustrate, Figure \ref{fig:dhi_stock_projection} shows a predicted forecast generated by the N-HiTS model compared to the actual adjprc. Although the model does not capture the small fluctuations of the actual adjprc, the overall trend of the stock is learned. The lack of small fluctuations in predictions can be explained by the N-HiTS model being trained on 360 different stocks, rather than a single stock, improving the generalizability of the forecasting model while sacrificing learning the local noise of the stock.\\

\section{Threat Model}
In this paper, it is assumed that ML/AI forecasting models for sensitive data, such as stock data, are regularly used by consumers and in production. This is an easily foreseeable assumption given the recent advancements of ML and AI, and the increased use of platforms such as Hugging Face. The attacker aims to alter existing stock data used as input for the forecasting model in hopes to influence the market. All attacks are performed in a white-box setting, meaning the attacker has access to the forecasting model's gradients so they can generate stronger attacks. Furthermore, this paper argues that the discussion between gradient-based and black-box methods can be avoided with other cyber-attacking tools such as malware. If a malware is present on a victim's device, any calls to prevent gradients, like torch.no\_grad(), can easily be avoided and removed. Finally, current ML security is focused on making the model robust, researching different ways to prevent model theft and poisoning attack \cite{ml_documentation:2025}. However, current research does not focus on the vulnerabilities of the ML interface once the model has already been deployed, where malware attacks are more likely to occur \cite{ml_documentation:2025}.

\section{Adversarial Attack Methods}
For the remainder of the paper, the adversarial input will be denoted as $x_{adv}$.
\subsection{Baseline White Box Attacks}
Several white-box adversarial attacks on images and time-series data, drawn from related literature, are considered as baselines to the proposed slope-based methods. These baseline methods include the Fast Gradient Sign Method (FGSM), Basic Iterative Method (BIM), Momentum Iterative Fast Gradient Sign Method (MI-FGSM), Stealthy Iterative Method (SIM), and the Targeted Iterative Method (TIM). Explanations of each attack method is provided in the Appendix.

\subsection{Slope Attacks}
A limitation of current targeted adversarial attack methods (TIM and C\&W) on time-series data is the lack of fine control over temporal characteristics without the cumbersome process of choosing a target $Y^*$ for each attack. For example, if the TIM adjusts the target forecast with a direction and a margin then the attack only controls how much the model over or under-predicts. However, in many sensitive domains like financial time-series data, adjusting the prediction up or down has little reward for the attacker as the general trend of the forecast remains the same. While the trend can be adjusted by defining a target forecast $Y^*$ with a specific slope, it is unrealistic for an attacker to create their own target for different inputs. Instead, a more applicable approach is to embed the desired slope inside the loss function we wish to minimize. Therefore, two slope-based attacks are introduced, the General Slope Attack and the Least-Squares Slope attack, as a means of effectively altering the temporal dynamics. 

\subsubsection{General Slope Attack}
In the simplest case, the attacker would want to only affect the endpoints of a prediction, as this might show a late positive surge or decline affecting the victim's desire to purchase a stock. Following the structure of Algorithm \ref{alg:slope_algo}, the General Slope Attack (GSA) uses the discrete formula to find the slope between two points. That is,
\begin{align*}
    m = \frac{y_2 - y_1}{x_2 - x_1}
\end{align*}
where $(x_1, y_1)$ and $(x_2, y_2)$ are the endpoints of the prediction. Then, the objective function requires the slope $m$, target direction $t \in \{-1, 0, 1\}$, and hyperparameters $c$ and $d$ and calculates the following loss:
\begin{equation} \label{eq:slope_loss}
    loss = 
        \begin{cases}
            ce^{-tdm} & \text{if $t \in \{-1, 1\}$  }\\
            cm^2 & \text{if $t = 0$}\\
        \end{cases}
\end{equation}
Based on the trend the attacker wishes the prediction to move in, the objection function heavily penalizes slopes in the incorrect direction. Furthermore, given that only the endpoints are used when calculating the slope, the data points in the middle of a time-series are largely unaffected, improving the stealthiness of the attack.

\subsubsection{Least-Squares Slope Attack}
While the GSA attack aims to change the slope of the time-series forecast near the endpoints of the prediction, the nature of the slope calculation prevents the overall trend from changing. This might not be useful for an attacker if they manipulate a downward moving stock, since changing the endpoints would not affect the downward trend, preventing any victim from buying the stock. Therefore, the slope calculation of the Least-Squares Slope Attack (LSSA) aims to alter the overall trend of the forecast.\\

Inspired by Least Squares Linear Regression, the new calculation for the slope would be the weight term in the closed form solution for Least-Squares Linear Regression. Specifically,
\begin{align*}
    m = \frac{\Sigma_{i=0}^{N} (x - \tilde{x})(y - \tilde{y})}{(x - \tilde{x})^2}
\end{align*}
The new slope calculation is simply the slope of the line of best fit, thus improving the loss would optimize the entire trend of the prediction, rather than just the endpoints. The new slope is also inputted into the loss function seen in Equation \ref{eq:slope_loss}.\\

\begin{algorithm}
    \textbf{Input:} Stock Forecasting model \textbf{model}, adjusted price \textbf{adjprc}, number of iterations $\textbf{iter} \in \mathbb{R}$, maximum perturbation $\boldsymbol{\epsilon} \in \mathbb{R}_{\ge 0}$, the step size $\boldsymbol{\alpha} = 1.5 \cdot \boldsymbol{\epsilon} / \textbf{iter} \in \mathbb{R}_{\ge 0}$, target direction $\mathbf{t} \in \{-1, 0, 1\}$, and scalars $\mathbf{c, d} \in \mathbb{R}$
    \begin{algorithmic}[1]
    \caption{Generic slope attack for financial time-series data}
    \label{alg:slope_algo}
    \State $x_{adv}$ $\gets$ adjprc
    \For{i \textbf{in} range(iter)}
        \State $x_{adv} \gets \text{requires grad}$
        \State $\text{model} \gets \text{zero grad}$
        \State $\text{features} \gets getFeatures(x_{adj})$
        \State $\text{output} \gets model(\text{features})$
        \State $m \gets SlopeMethod(\text{output})$
        \If {$t \in \{-1, 1\}$}
            \State $\text{loss} \gets ce^{-tdm}$
        \Else 
            \State $\text{loss} \gets cm^2$
        \EndIf
        \State loss.backward()
        \State \text{with} \text{no grad}:
            \State $\text{\quad \quad} \eta \gets \alpha \cdot sign(\nabla x_{adv})$
            \State $\text{\quad \quad} x_{adv} \gets x_{adv} - \eta$
            \State $\text{\quad \quad} x_{adv} \gets clamp(x_{adv}, \text{adjprc} -\epsilon, \text{adjprc} + \epsilon)$
        \State $\text{detach } x_{adv}$
    \EndFor
    \State \textbf{return} $x_{adv}$
    \end{algorithmic}
\end{algorithm}
\FloatBarrier

Note that the GSA and LSSA could easily be converted into C\&W-type attacks, where the adversarial loss is replaced with one of the slope-based objective functions. Sample C\&W versions of the slope-based attacks are provided in the Appendix.

\subsection{Generative Adversarial Network Based Attacks}
\subsubsection{Generative Adversarial Networks}
GANs, originally created by I. Goodfellow et al., describe a neural network with two submodels, a generator and discriminator \cite{gans:2014}. The generator $G$, aims to create synthetic data by learning the distribution of the real data, while the discriminator $D$, aims to distinguish the real from the synthetic data \cite{gans:2014}. The objective function of a vanilla GAN is: 
\begin{align*}
    \min_G \max_D \mathbb{E}_{x\sim p_{data}(x)}[\log D(x)] + \mathbb{E}_{z \sim p_{z}(z)}[(\log(1- D(G(z)))] \text{\quad \quad \cite{gans:2014}}
\end{align*}
Essentially, the discriminator aims to maximize performance when classifying real data, while the generator aims to maximize the performance of its synthetic data being labeled as real by the discriminator.\\

\subsubsection{Adversarial GANs}
GANs have previously been used to create adversarial attacks, as they are able to make believable inputs by learning the underlying characteristics of its dataset \cite{maggan:2020}. As an attacker, the main goal of an adversarial attack is to avoid detection; thus, an attacker can leverage the generated realism of GANs to make believable adversarial attacks. Figure \ref{fig:adv_gan_architecture} displays the architecture of an Adversarial GAN (A-GAN), where a second critic is attached to the GAN architecture, which is the N-HiTS model we wish to fool. \\

The A-GAN was trained with stock data extracted similar to the N-HiTS model, specifically using the stock with the ticker $A$; however, due to its stationary properties, the current implementation generates 99 days of log returns rather than adjprc. Random continuous intervals of 99 days were selected for each sample of training data. Furthermore, min-max scaling was applied on the training data. Although the N-HiTS model works in price space, experiments have shown that generating sequences of adjprc leads to worse GAN performance. \\

The noise vector is initialized to be the same size as the desired synthetic data (99 days) and is passed to the generator, made of a 4-layered Temporal Convolutional Network (TCN). The TCN architecture was chosen to help capture long term dependencies. The critic is a hybrid architecture with 5 layers, and is made up of alternating TCN and Gated Recurrent Network (GRU) blocks (3 TCN blocks, 2 GRU blocks) with a similar focus on long term dependencies. For both sub-models, a final linear layer was used to reduce the output dimension to 1.\\

A Conditional Wasserstein GAN (C-WGAN) was used for the A-GAN, conditioned with the corresponding 99 days of log returns. The condition is then concatenated with the noise vector in the feature dimension for both the generator and the critic. Thus, the critic learns that the provided data should be extremely similar to the condition, while the generator learns not to deviate too far from its condition. Conditioning with previous information was also experimented with, however, the performance was much worse. Therefore, more work is needed to improve the training and stability of the A-GAN architecture.\\

To incorporate the adversarial loss, the generator's output was converted from log returns to adjprc using the initial price from the condition. Then, after the min-max scaling was reversed, the adjprc was fed into the N-HiTS model and the LSSA was performed. Specifically, the loss of the generator is defined as:
\begin{align*}
    L_g = \mathbb{E}_{\tilde{x} \sim P_{g}}[D(\tilde{x})] + \alpha \cdot LSSA(\textit{N-HiTS}(\tilde{x}))
\end{align*}
where $\alpha$ is a scaling factor to balance the adversarial loss and the original generator loss. Note that the GSA is not used since the gradients would only flow through the endpoints of the prediction, making it not suitable for GAN training. The original critic is trained as normal, with gradient penalties only added to the critic loss 60\% of the time to improve training speed. Therefore, the final A-GAN loss is defined as:
\begin{align*}
    L = L_g - \mathbb{E}_{x \sim P_{r}}[D(x)] + \lambda \cdot \mathbb{E}_{\hat{x} \sim P_{\hat{x}}} [(||\nabla_{\hat{x}} D(\hat{x})|| - 1)^2]
\end{align*}

However, adding the second critic to the GAN architecture was not a trivial task. In order to improve stability and training performance, the A-GAN was trained in batches of 50 epochs with the occasional increase in scaling values $\alpha$ for the adversarial loss. Then transfer learning was used for each training block, where the model was trained five times, with $\alpha$ values [0.25, 0.25, 0.3, 0.35, 0.35]. Experiments have shown that increasing the adversarial scale too much would cause unrealistic synthetic data. All other hyperparameters were determined experimentally and are defined in Table \ref{tab:wgan_h_params} in the Appendix. Further work is necessary to improve the training and stability of the A-GAN.\\

Several versions of the GAN architecture were experimented with, such as a Deep Convolutional GAN (DCGAN) and an unconditioned Wasserstein GAN (WGAN), before a Conditional WGAN (C-WGAN) was chosen. Explanations of the DCGAN and WGAN are given in the Appendix.

\section{Results}
\subsection{Iterative Attacks}
\begin{figure}[ht]
\centering
\begin{subfigure}{0.95\textwidth}
    \includegraphics[width=\textwidth]{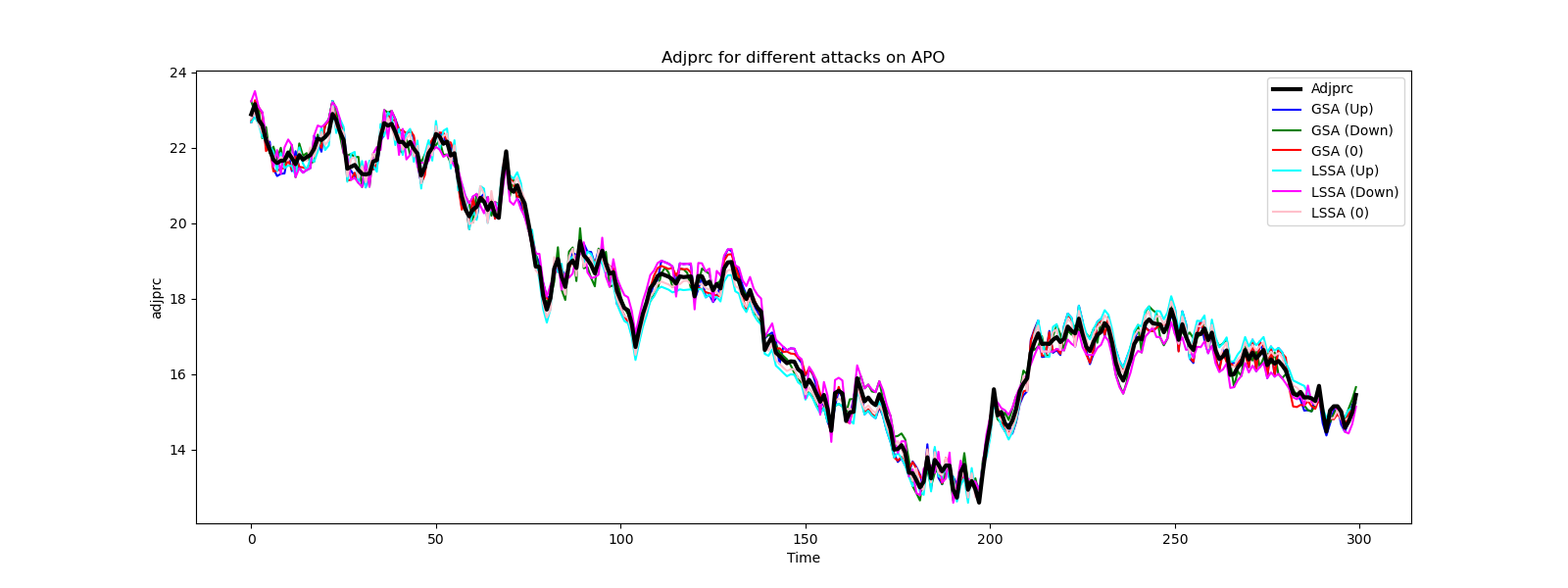}
    \caption{APO Adjprc}
    \label{fig:apo_adjprc}
\end{subfigure}
\begin{subfigure}{0.95\textwidth}
    \includegraphics[width=\textwidth]{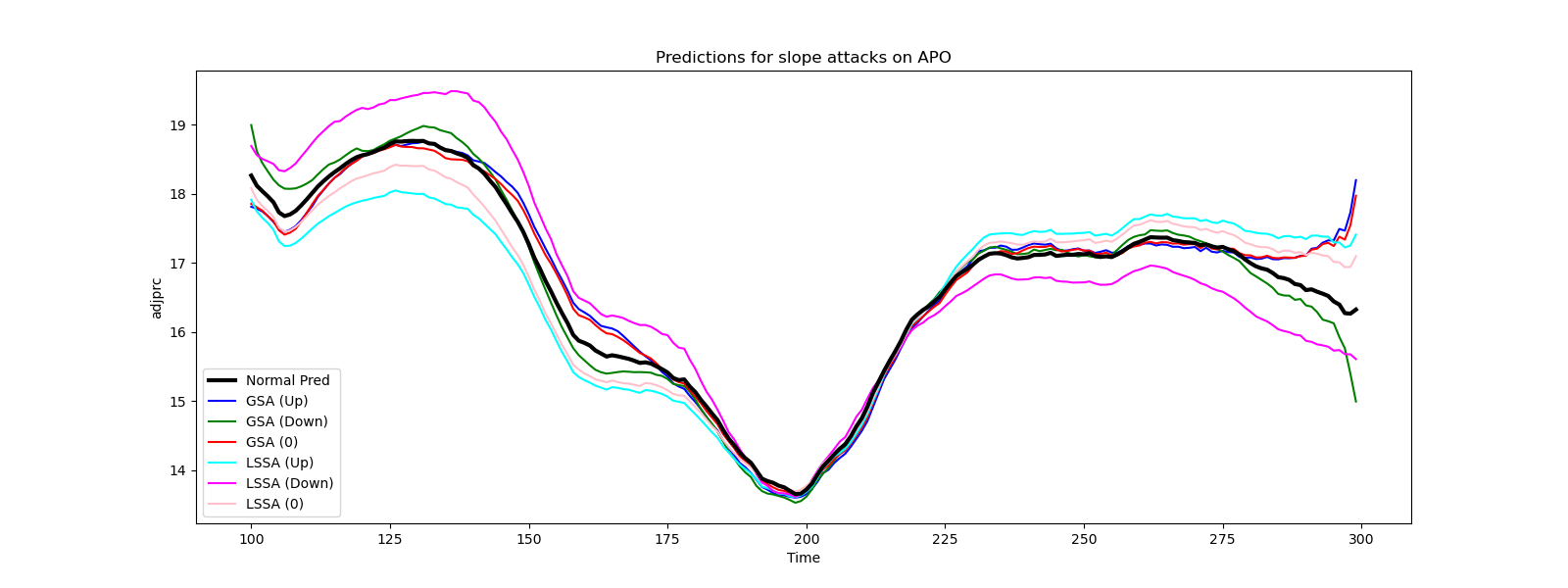}
    \caption{APO Attack}
    \label{fig:apo_attack}
\end{subfigure}
\caption{Slope attacks, and their perturbed adjprc, for the first 300 days of the recording for stock $APO$.}
\label{fig:apo_attack_example}
\end{figure}

Given that the N-HiTS model requires a look-back period of 100 days, a segment of 300 days was chosen to give the N-HiTS model ample moving windows, improving the overall performance and robustness. Attacks were performed on the first 300 days of each recording in the test set, due to the low MAE for the normal forecast. Predictions made on the unperturbed adjprc had an average MAE of 2.15. Furthermore, a relative epsilon was chosen for each stock to avoid issues with different scales. The epsilon was calculated by taking the median of the adjprc and multiplying it by a small percentage, denoted $\epsilon$ \%. Table \ref{tab:attack_metrics} shows the results for the GSA and LSSA compared to other baseline methods, with the $\epsilon$ \% being 2\%, meaning that the daily price can change by at most 2\%. The small magnitudes in the calculated slopes for each attack are caused by the large prediction window of 200 days normalizing the slopes to very small values. Nevertheless, the GSA is able to double the general slope of the normal, unperturbed prediction for the upward direction, in addition to decreasing the general slope to below zero for the downward direction. This shows the ability of the GSA to cause strong shifts near the endpoints which is further exemplified in Figure \ref{fig:apo_attack_example}. The same trend can be seen with the LSSA attack, where the LS slope is doubled in the upward direction, while being brought down to 0 in the positive direction. Note that the downward versions of the slope attacks are able to outperform the zero-slope forms, most likely due to the quadratic penality not being strong enough compared to the exponential.

\begin{table}[ht]
    \centering
    \caption{Average metrics for different attack methods performed on the first 300 days of each recording, with $\epsilon = 2\% \cdot median(adjprc)$. The best metrics are bolded.}
    \begin{tabular}{c|ccc|cc}
        \hline
        \textit{Attack} & \textit{MAE} & \textit{RMSE} & \textit{MAPE} & \textit{Gen. Slope} & \textit{LS Slope}\\
        \hline
        Normal & 2.15 & 2.72 & 3.82$\times 10^{-2}$ & 3.37$\times 10^{-2}$ & 2.22$\times 10^{-2}$\\
        \hline
        FGSM & 2.57 & 3.21 & 4.51$\times 10^{-2}$ & 3.22$\times 10^{-2}$ & 2.34$\times 10^{-2}$\\
        BIM & \textbf{3.38} & \textbf{3.99} & \textbf{5.68}$\mathbf{\times 10^{-2}}$ & 3.48$\times 10^{-2}$ & 2.39$\times 10^{-2}$\\
        MI-FGSM & 3.37 & \textbf{3.99} & 5.67$\times 10^{-2}$ & 3.44$\times 10^{-2}$ & 2.39$\times 10^{-2}$\\
        SIM & 2.57 & 3.08 & 4.29$\times 10^{-2}$ & 3.37$\times 10^{-2}$ & 2.23$\times 10^{-2}$\\
        TIM (Up) & 2.49 & 3.21 & 4.52$\times 10^{-2}$ & 3.72$\times 10^{-2}$ & 2.00$\times 10^{-2}$\\
        TIM (Down) & 2.74 & 3.26 & 4.44$\times 10^{-2}$ & 3.32$\times 10^{-2}$ & 2.51$\times 10^{-2}$\\
        \hline
        \textit{GSA (Up)} & 2.26 &  2.88 & 4.03$\times 10^{-2}$ & \textbf{6.76}$\mathbf{\times 10^{-2}}$ & 2.77$\times 10^{-2}$\\
        \textit{GSA (Down)} & 2.23 & 2.83 & 3.89$\times 10^{-2}$ & \textbf{-1.68}$\mathbf{\times 10^{-4}}$ & 1.75$\times 10^{-2}$\\
        \textit{GSA (0)} & 2.30 & 2.93 & 4.01$\times 10^{-2}$ & 1.80$\times 10^{-2}$ & 2.00$\times 10^{-2}$\\
        \hline
        \textit{LSSA (Up)} & 2.49 & 3.10 & 4.26$\times 10^{-2}$ & 5.38$\times 10^{-2}$ & \textbf{4.96}$\mathbf{\times 10^{-2}}$\\
        \textit{LSSA (Down)} & 2.71 & 3.33 & 4.63$\times 10^{-2}$ & 1.56$\times 10^{-2}$ & \textbf{-5.04}$\mathbf{\times 10^{-3}}$\\
        \textit{LSSA (0)} & 2.68 & 3.31 & 4.55$\times 10^{-2}$ & 2.82$\times 10^{-2}$ & 1.29$\times 10^{-2}$\\
        \hline
    \end{tabular}
    \label{tab:attack_metrics}
\end{table}
Unsurprisingly, the slope-based methods improve with larger sizes of $\epsilon$ \%, as shown in Figure \ref{tab:epsilon_experiment}. However, a small $\epsilon$ \% in 0.5\% is able to produce a meaningful general slope improvement compared to the normal prediction for both attack methods, showing that the attacks are effective with minimal perturbation.
\begin{table}[ht]
    \centering
    \caption{Average slope values for varying relative percentages. Only the upward version of the attacks are considered since all other attacks follow the same trend.}
    \begin{tabular}{c|ccc}
    \hline
    \textit{Attack} & \textit{$\epsilon$ \%} & \textit{Gen. Slope} & \textit{LS Slope}  \\
    \hline
    Normal & 0.0 & 3.37$\times 10^{-2}$ & 2.22$\times 10^{-2}$\\
    \hline
    \multirow{8}{*}{GSA (Up)} & 0.5  & 4.37$\times 10^{-2}$ & 2.41$\times 10^{-2}$\\
                         & 1.0  & 5.27$\times 10^{-2}$ & 2.57$\times 10^{-2}$\\
                         & 1.5  & 6.06$\times 10^{-2}$ & 2.70$\times 10^{-2}$ \\
                         & 2.0  & 6.76$\times 10^{-2}$ & 2.77$\times 10^{-2}$ \\
                         & 2.5 & 7.39$\times 10^{-2}$ & 2.84$\times 10^{-2}$ \\
                         & 3.0 & 7.93$\times 10^{-2}$ & 2.90$\times 10^{-2}$  \\ 
                         & 3.5 & 8.46$\times 10^{-2}$ & 2.94$\times 10^{-2}$ \\
                         & 4.0 & 8.93$\times 10^{-2}$ & 3.02$\times 10^{-2}$ \\
    \hline
    \multirow{8}{*}{LSSA (Up)}  & 0.5 & 3.94$\times 10^{-2}$ & 2.97$\times 10^{-2}$  \\
                         & 1.0 & 4.45$\times 10^{-2}$ & 3.67$\times 10^{-2}$ \\
                         & 1.5  & 4.95$\times 10^{-2}$ & 4.33$\times 10^{-2}$ \\
                         & 2.0  & 5.38$\times 10^{-2}$ &  4.96$\times 10^{-2}$\\
                         & 2.5  & 5.79$\times 10^{-2}$ & 5.55$\times 10^{-2}$ \\
                         & 3.0 & 6.18$\times 10^{-2}$ & 6.12$\times 10^{-2}$\\ 
                         & 3.5 & 6.53$\times 10^{-2}$ & 6.67$\times 10^{-2}$ \\
                         & 4.0 & 6.90$\times 10^{-2}$ & 7.22$\times 10^{-2}$  \\
    \hline
    \end{tabular}
    \label{tab:epsilon_experiment}
\end{table}
\FloatBarrier
\FloatBarrier
\subsection{A-GAN Attacks}
\begin{figure}[ht]
\centering
\begin{subfigure}{0.45\textwidth}
    \includegraphics[width=\textwidth]{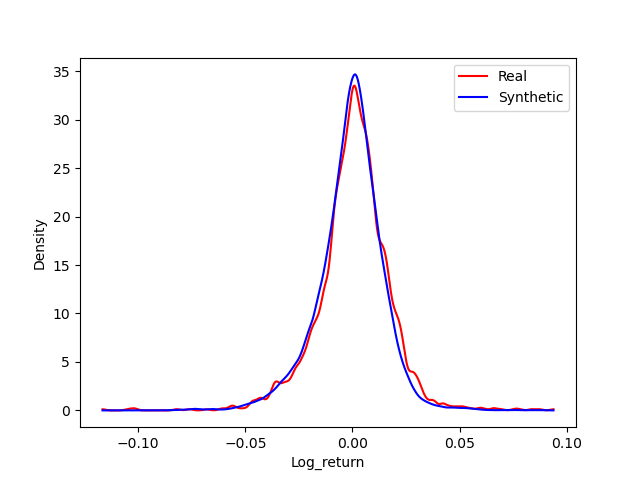}
    \caption{A-GAN KDE}
    \label{fig:agan_kde}
\end{subfigure}
\begin{subfigure}{0.45\textwidth}
    \includegraphics[width=\textwidth]{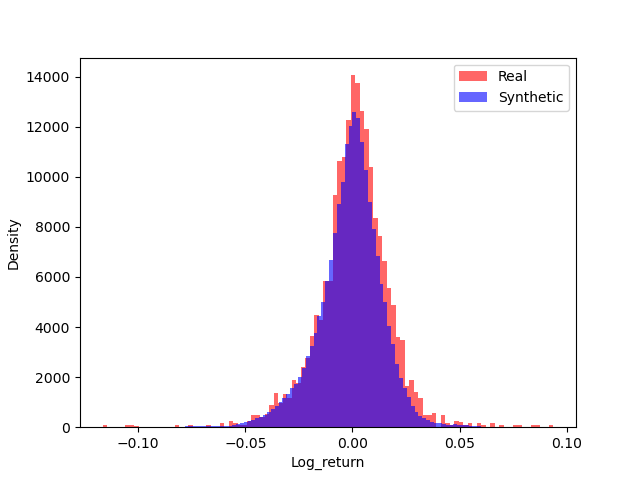}
    \caption{A-GAN Histogram}
    \label{fig:agan-hist}
\end{subfigure}
\caption{KDE and Histogram Plots of the distributions between the real and synthetic data generated by the A-GAN, after sampling 2000 intervals.}
\label{fig:a_gan_hist_kde}
\end{figure}

To evaluate the A-GAN, 2000 log returns (99 day intervals) from the real stock were sampled and used as conditions to generate 2000 intervals of synthetic log returns. From the log returns, the mean ($\mu$), standard deviation ($\sigma$), Inter-quartile range (IQR), skew, and raw kurtosis were measured. In addition, the Maximum Mean Discrepancy (MMD) was measured between the real and fake samples, with the gamma value calculated by the median heuristic \cite{medical_gans:2017}. As illustrated in Figure \ref{tab:gan_slope_metrics}, the A-GAN is able to vastly increase the trend of the prediction of the N-HiTS model. Unsurprisingly, the general slope essentially remains unchanged as this was not what the A-GAN was optimized for. Furthermore, Figure \ref{tab:gan_metrics} shows how the moment-based statistics of the synthetic data generated by the A-GAN is extremely similar to the corresponding real data, except for the mean. The vastly different mean is a result of the GAN generating results in log-space, while the N-HiTS model works in price-space. Since the conversion from log space to price space accumulates error, slight shifts in the mean can have drastic affects on the converted adjprc. Therefore, when incorporating the adversarial loss, the A-GAN learns that in order to minimize the slope error, it is enough to modify the mean of the log returns. However, the extremely small MMD shows that the synthetic data is faithfully similar to the real data. Furthermore, the KDE and Histogram plots in Figure \ref{fig:a_gan_hist_kde} solidify the similarity between the real and synthetic data.

\begin{table}[ht]
    \centering
    \caption{Slope comparisons based on sampling the real and generated data 2000 times.}
    \vspace{-3mm}
    \begin{tabular}{c|cc}
        \hline
        \textit{Data} & \textit{General Slope} & \textit{LS Slope}\\
        \hline
        Real & 0.999 & -2.41$\times 10^{-3}$ \\
        A-GAN & 1.050 & 2.17$\times 10^{-1}$\\
        \hline
    \end{tabular}
    \label{tab:gan_slope_metrics}
\end{table}

\begin{table}[ht]
    \centering
    \caption{Statistical metrics based on sampling the real and generated data 2000 times.}
    \vspace{-3mm}
    \begin{tabular}{c|cccccc}
        \hline
        \textit{Data} & \textit{$\mu$} & \textit{$\sigma$} & \textit{IQR} & \textit{Skew} & \textit{Kurtosis} & \textit{MMD}\\
        \hline
        Real & 5.71$\times 10^{-4}$ & 1.63$\times 10^{-2}$ & 1.82$\times 10^{-2}$ & -3.81$\times 10^{-1}$ & 5.74 & 0 \\
        A-GAN & -9.85$\times 10^{-4}$ & 1.48$\times 10^{-2}$ &  1.72$\times 10^{-2}$ & -4.26$\times 10^{-1}$ & 4.31 & 1.20$\times 10^{-4}$\\
        \hline
    \end{tabular}
    \label{tab:gan_metrics}
\end{table}

\begin{figure}[H]
    \centering
    \includegraphics[scale=0.55]{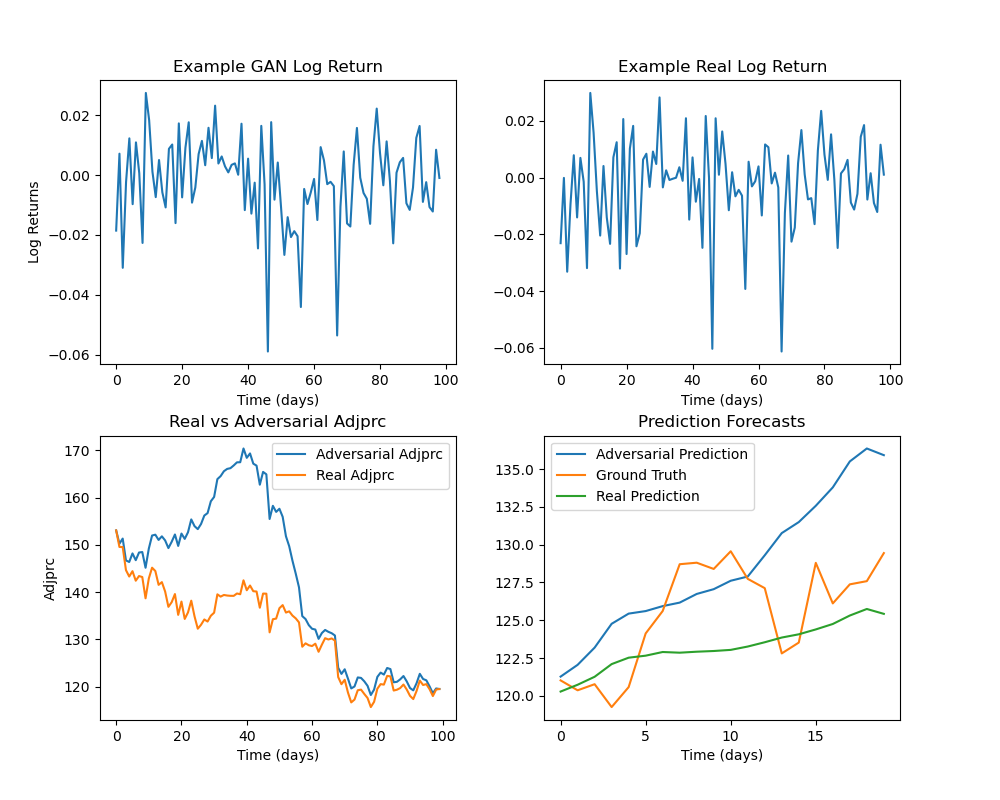}
    \vspace{-2mm}
    \caption{Example A-GAN output generated from a random interval.}
    \label{fig:a_gan_output}
\end{figure}

\FloatBarrier
Figure \ref{fig:a_gan_output} provides an example of a log return generated by the A-GAN. The prediction forecast the ability for the GAN to increase the slope of the prediction, however the slight differences in the log returns lead to large differences in certain areas in the adjprc. Further examples of the A-GAN outputs are shown in the Appendix.

\section{Defense Mechanisms}
There are numerous examples of adversarial defense strategies in related literature; however, the main defense used is adversarial training \cite{pialla:2025}. Adversarial training is the standard practice of making a model more robust to adversarial attacks. Adversarial training involves including the adversarial examples with the real data in the training set to effectively train the model to correctly classify similar inputs regardless of noise \cite{pialla:2025}. However, adversarial training was not used for the N-HiTS model due to infeasibility. Given that the N-HiTS model uses the adjprc as a feature, in addition to generating predictions in a rolling window, there is no feasible way to alternate 100 days of adversarial data with its ground truth 20 day forecast without creating separate files for each 100 day segment. Given that the training data used contains over 300 recordings made up of 10 years of stock data, creating this many files is not feasible, illustrating a limitation of training an N-HiTS model in this manner.\\

As an alternative, a developer can create a discriminator aimed to distinguish adversarial and real data \cite{pialla:2025}. During inference, the discriminator is run and returns the likelihood of the input being altered. If the discriminator detects that the given input has been tampered with, predictions are not made.\\ 

To test the stealthiness of the new attacks, a 4-layered CNN was trained on varying sets of adversarial methods. Hyperparameters of the discriminator are given in the Appendix. As shown by Figure \ref{fig:discrim_cm}, the CNN is unable to identify adversarial inputs from real ones, with the model trained on GSA achieving an accuracy of 52.08, specificity of 27.78 and $\kappa$ of 4.17 and the model trained on LSSA obtaining an accuracy of 56.25, specificity of 26.40 and $\kappa$ of 12.50. Thus, these new iterative slope-based methods are able to remain stealthy, bypassing standard protection against adversarial attacks. Although the iterative methods are able to remain undetected, the mode collapse of the A-GAN causes the synthetic data to easily be identified, as the CNN is able to obtain an accuracy of 92.83, specificity of 92.66 and $\kappa$ of 85.67. \\

\begin{figure}[ht]
\centering
\begin{subfigure}{0.33\textwidth}
    \includegraphics[width=\textwidth]{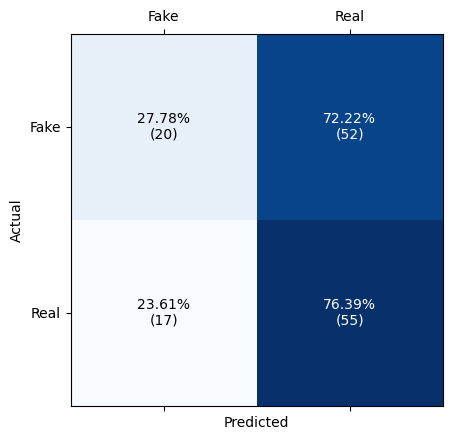}
    \caption{GSA}
    \label{fig:gsa_discrim}
\end{subfigure}
\begin{subfigure}{0.33\textwidth}
    \includegraphics[width=\textwidth]{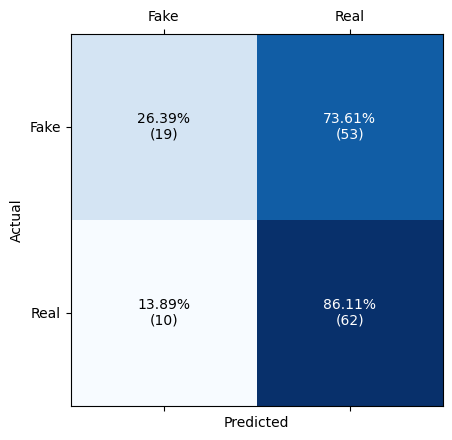}
    \caption{LSSA}
    \label{fig:lssa_discrim}
\end{subfigure}
\begin{subfigure}{0.33\textwidth}
    \includegraphics[width=\textwidth]{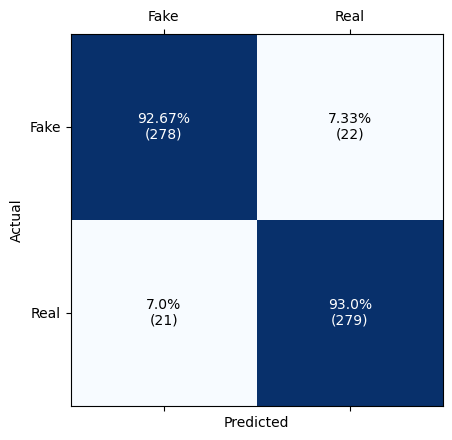}
    \caption{A-GAN}
    \label{fig:agan_discrim}
\end{subfigure}
\caption{Confusion matrices for the discriminator trained to identify GSA and LSSA attacks with $\epsilon$ \% $= 2$}
\label{fig:discrim_cm}
\end{figure}
\FloatBarrier

\section{Trojan Malware for Model Interface}

Current adversarial attacks and ML security research typically focus on securing the model rather than the entire ML pipeline/interface. For example, many studies focus on preventing attacks on the model such as model stealing, model weight alteration, or poisoning attacks \cite{ml_documentation:2025}. Developers typically assume that the deployed system to use the model is safe; however, these systems can be easily exploited.\\

A common workflow for deployment may be to have a project package/library that stores the saved model and all files required to make a prediction. In such cases, the prediction directory is thought to be safe and secure, yet can be easily manipulated with existing cyber-attack methods. For example, in Python, the \_\_init\_\_.py file is run whenever something inside the package is imported. However, if the \_\_init\_\_.py was tampered with beforehand, the attacker can modify any code inside the project directory. Thus, the attacker can easily inject an adversarial attack, bypassing any filtering/detection systems the model may have. Furthermore, the attacker can remove any calls to disable gradients, allowing for stronger white-box attacks to occur. An example of a malware is shown in Algorithm \ref{alg:sample_malware}, illustrating the relative ease in which an attacker can create a self-cleaning attack that is run during every call to the project package. The attacker may need some white-box knowledge, such as the library used to develop the model (PyTorch vs Tensorflow); however, the creation of the malware is not infeasible.
\begin{algorithm}
    \begin{algorithmic}[1]
    \caption{Sample payload for \_\_init\_\_.py file}
    \label{alg:sample_malware}
    \State adversarialString $\gets$ "Adversarial Attack Code"
    \State originalFile $\gets$ fileWithModelCall.read()
    \State originalFileLines $\gets$ fileWithModelCall.readlines()
    \State f $\gets$ open(fileWithModelCall)
    \For{line in originalFileLines}
        \If {line has a model call}
            \State write adversarialString
        \EndIf
        \If {line has no\_grad call}
            \State for each line with larger indent, remove indent and write
        \Else
            \State write line
        \EndIf
    \EndFor
    \State at exit, write originalFile to fileWithModelCall
    \end{algorithmic}
\end{algorithm}
\FloatBarrier

There are two scenarios where such an attack could occur, an external and an internal attacker. An external attacker is someone who did not develop the model or project package, while an internal attacker is someone who was part of the development team. If the attack was carried out externally, then the malware described in Algorithm \ref{alg:sample_malware} could be mitigated by hashing the project directory with the installer verifying the hash. If the hash provided by the developers does not match the project directory, then there has been some form of tampering. However, if the attack was performed internally, then there are very few methods to mitigate the attack, as the directory with the malware will be part of the hash. The only clear way to prevent such an attack would be a code review. With the increasing number of ML/AI developers, in addition to the countless models being published, it is not unrealistic for ML/AI models to be downloaded like apps and libraries. Similar to how there are nefarious app developers who remain anonymous, certain ML/AI engineers may also have immoral intentions, illustrating the importance of maintaining good security practices over the entire pipeline.

\section{Discussion and Conclusion}
\subsection{Key Findings}
In this study, slope-based attacks designed for financial time-series data were investigated to determine if they can strike a balance between generating realistic adversarial inputs while also having a strong affect on the victim model. Not only do the GSA and LSSA show a drastic effect on a state-of-the-art N-HiTS architecture trained for financial forecasting, they also have the ability to covertly bypass standard security measures. Furthermore, incorporating the slope-based attack into the GAN architecture has shown that GANs are able to effectively generate synthetic adversarial examples.\\

Although the A-GAN has shown to successfully increase the slope of the victim model's forecast, the A-GAN suffers from mode collapse, which occurs when the generator creates data with limited diversity \cite{mode_collapse:2024}. The extra examples in the Appendix illustrate this phenomenon, where each adjprc generated has a large initial increase followed by a steep drop-off. Then the increase in the N-HiTS prediction occurs due to the assumption that the stock will bounce back. As a result, each generated output is extremely similar, making it easier for the discriminator to catch the adversarial examples generated by the A-GAN. However, this does provide an insight on the behaviour of the N-HiTS model. \\

Finally, this paper illustrates the importance of ML/AI security throughout the entire pipeline and interface as it may also be viable to attack areas such as the model library with existing cyber-attack methods. The sample malware, shown in this paper, demonstrates how an attacker could inject adversarial attacks that use gradient information, allowing for a stronger and more consequential attack. Therefore, it is imperative for more research to focus on securing the entire machine learning pipeline, given that no model in any domain is truly safe against adversarial attacks. 

\subsection{Limitations and Future Work}
This paper has several limitations. First, in order to solidify and prove the general effectiveness of the attack methods, several other ML models, like CNNs and LSTMs, should have been developed and used as the victim models, similar the study by Shen and Li \cite{tar_and_stealthy:2025}. In addition, even though adversarial training is not a trivial or necessarily feasible task for the model developed in this study, it should still be performed to determine the effectiveness of the standard defense practice. Adversarial training could be performed by training a new model which does not rely on adjprc as a feature. Furthermore, to prove the applicability of the slope-based attacks on time-series data, these attacks should be implemented in other time-series domains that use forecasting models, such as traffic and electricity usage. Finally, the training stability and performance of the A-GAN remains a significant issue. Future work will explore methods to improve stability, in addition to preventing mode collapse. 

\section*{Acknowledgements}
Thank you to Professor Irene Huang, at the University of Toronto Scarborough, for supervising me during this CSCD94 undergraduate research project.

\newpage
\section{Appendix}
\subsection{Baseline Adversarial Attacks}
Most iterative adversarial attacks follow a similar pattern, as described in Algorithm \ref{alg:generic_adv}, where the attacker would iteratively probe the model with its adversarial input, $x_{adv}$, compute a loss, and update $x_{adv}$ with respect to the gradient of the loss. However, each baseline attack either updates $x_{adv}$ or computes the loss differently. Note that for this study, the L1 Loss was used in Algorithm \ref{alg:generic_adv}, as the goal was to maximize the MAE, however any regression loss function would work.
\begin{algorithm}
    \textbf{Input:} Stock Forecasting model \textbf{model}, adjusted price \textbf{adjprc}, number of iterations $\textbf{iter} \in \mathbb{R}$, maximum perturbation $\boldsymbol{\epsilon} \in \mathbb{R}_{\ge 0}$, and the step size $\boldsymbol{\alpha} = 1.5 \cdot \boldsymbol{\epsilon} / \textbf{iter} \in \mathbb{R}_{\ge 0}$
    \begin{algorithmic}[1]
    \caption{Generic Iterative Adversarial Attack Algorithm on the N-HiTS model}
    \label{alg:generic_adv}
    \State $x_{adv}$ $\gets$ adjprc
    \For{i \textbf{in} range(iter)}
        \State $x_{adv} \gets \text{requires grad}$
        \State $\text{model} \gets \text{zero grad}$
        \State $\text{features} \gets getFeatures(x_{adj})$
        \State $\text{output} \gets model(\text{features})$
        \State $\text{loss} \gets LossFunction\text{(predictions, adjprc)}$
        \State loss.backward()
        \State \text{with} \text{no grad}:
            \State $\text{\quad \quad} x_{adv} \gets AttackMethod(x_{adv}, \alpha)$
        \State $\text{detach } x_{adv}$
    \EndFor
    \State \textbf{return} $x_{adv}$
    \end{algorithmic}
\end{algorithm}
\FloatBarrier
\subsubsection{FGSM}
The Fast Gradient Sign Method (FGSM) is the simplest way to create an adversarial attack. The FGSM involves computing the loss with respect to the forecasting predictions and the ground truth label (adjprc) then, using the sign of the gradients, $x_{adv}$ is adjusted to maximize the loss \cite{fgsm:2023}. That is:
\begin{align*}
    x_{adv} & = x + \alpha \cdot sign(\nabla_{x_{adv}} loss) \text{\quad \cite{fgsm:2023}}
\end{align*}

Note that the FGSM is done in a single step, thus, the step size $\alpha = \epsilon$. Normally, when a neural network performs gradient descent, the gradients determine the direction in which to move the weights. Since the gradient term is subtracted from the weights, the weights move opposite of the gradient's sign. However, the FGSM computes the loss with respect to each element in the time-series $x_{adv}$, and instead of subtracting the gradient, it is added, leading to the maximization of the loss \cite{fgsm:2023}.
\subsubsection{BIM}
The FGSM is a special case of the Basic Iterative Method (BIM), also called the Iterative Fast Gradient Sign Method (I-FGSM), where the number of iterations is 1 and the step size is $\epsilon$ \cite{mifgsm:2018}. The BIM applies the FGSM repeatedly for a desired number of iterations, but with a smaller step size $\alpha$ \cite{bim:2016}. The BIM follows the same generic algorithm shown in Algorithm \ref{alg:generic_adv}, initializing the adversarial attack $x_{adv}$ to the unperturbed adjprc, and then proceeds to iteratively update $x_{adv}$ by taking the sign of the gradient, and moving $x_{adv}$ in the direction that maximizes the loss. However, to ensure that $x_{adv}$ is in the $\epsilon$-neighborhood of the original time-series, $x_{adv}$ is clipped between adjprc - $\epsilon$ and adjprc + $\epsilon$ \cite{bim:2016}.

\subsubsection{MI-FGSM}
The Momentum Iterative Fast Gradient Sign Method (MI-FGSM) makes use of the momentum technique used to accelerate gradient descent algorithms \cite{mifgsm:2018}. By accumulating the direction of the loss function gradients across iterations, the momentum techniques allow gradient descent algorithms to avoid getting stuck in a local minima \cite{mifgsm:2018}. Let $g_t$ be the variable containing the gradients up to iteration t and let $\mu$ be the decay factor \cite{mifgsm:2018}. Initialize $g_0 = 0$, and $x_{adv}$ is updated by first computing $g_{t+1}$ as:
\begin{align*}
    g_{t+1} & = \mu \cdot g_t \cdot \frac{\nabla_{x_{adv}} loss}{||\nabla_{x_{adv}} loss||_1}
\end{align*}
then is moved in the direction of the $g_{t+1}$:
\begin{align*}
    x_{adv} & = x_{adv} + \alpha \cdot sign(g_{t+1})
\end{align*}
\cite{mifgsm:2018}. Finally, similar to BIM, $x_{adv}$ is clipped to stay within adjprc - $\epsilon$ and adjprc + $\epsilon$. The gradient of the loss is normalized by the 1-norm since each iteration may produce different magnitudes of gradients \cite{mifgsm:2018}. The current implementation of the MI-FGSM uses $\mu = 0.35$, which was determined experimentally.

\subsubsection{Stealthy Iterative Method}
An important factor to consider when designing an adversarial attack is that the perturbed time-series should be relatively undetectable. Although the size of $\epsilon$ controls the magnitude of the noise added to input, Z. Shen and Y. Li propose the Stealthy Iterative Method (SIM) to ensure the temporal characteristics of the time-series remain intact \cite{tar_and_stealthy:2025}. Z. Shen and Y. Li also extend the BIM by introducing cosine similarity comparisons between $x_{adv}$ and adjprc \cite{tar_and_stealthy:2025}. First, if the cosine similarity of adjprc and $x_{adv}$ is larger than the similarity between adjprc and adjprc + $\epsilon$, then $x_{adv}$ is kept as is, otherwise $x_{adv} \gets \text{adjprc} + \epsilon$ \cite{tar_and_stealthy:2025}. Then, if the cosine similarity between adjprc and $x_{adv}$ is larger than the cosine similarity between adjprc - $\epsilon$, $x_{adv}$ is kept as is, otherwise $x_{adv} \gets \text{adjprc} - \epsilon$ \cite{tar_and_stealthy:2025}.

\subsubsection{Targeted Iterative Method}
Although the goal for many adversarial attacks is to maximize the error of the model, in realistic scenarios the attacker would want to choose the direction in which the model fails. Thus, instead of simply maximizing the error, the Targeted Iterative Method (TIM) is another extension on the BIM which allows the attacker to add noise to adjprc, resulting in the forecast going in the direction that the attacker desires. There are two ways that the target can be calculated; first a new target sequence $Y^*$ can be defined by the attacker, or the algorithm takes in a direction $d$ and margin $\gamma$ and calculates a new target as:
\begin{align*}
    tar & = \text{adjprc} + d \cdot \gamma \text{\quad \quad \cite{tar_and_stealthy:2025}}
\end{align*}
Now, rather than modifying $x_{adv}$ to go in the direction that maximizes the loss between the predictions and adjprc, the loss between the predictions and $tar$ is computed instead \cite{tar_and_stealthy:2025}. Given that the goal is to minimize the error between the predictions and $tar$, for a given step size $\alpha$, $x_{adv}$ is updated as:
\begin{align*}
    x_{adv} = x_{adv} - \alpha \cdot sign(\nabla_{x_{adv}} loss) \text{\quad \quad  \cite{tar_and_stealthy:2025}}
\end{align*} 

\subsubsection{C\&W}
The Carlini and Wagner attack (C\&W) deviates from the general approach to adversarial attacks. Rather than iteratively modifying $x_{adv}$, a C\&W attack aims to find the optimum noise vector $\eta$ that is added to the adjprc \cite{pialla:2025}. Specifically, the C\&W attack aims to solve the following optimization problem:
\begin{align*}
    \min ||\eta||_2 + f(x_{adv} + \eta)
\end{align*}
where $f$ is a function that enforces an incorrect prediction \cite{pialla:2025}. Since this research focuses on a regressive task, $f$ was chosen to be the L1 Loss between the attack predictions and a target $tar$ defined in Section 3.5. In the following sections, the C\&W attack performed was targeted below the original adjprc.

\subsubsection{Baseline Attacks on Individual Stocks}
\begin{figure}[H]
    \centering
    \includegraphics[scale=0.35]{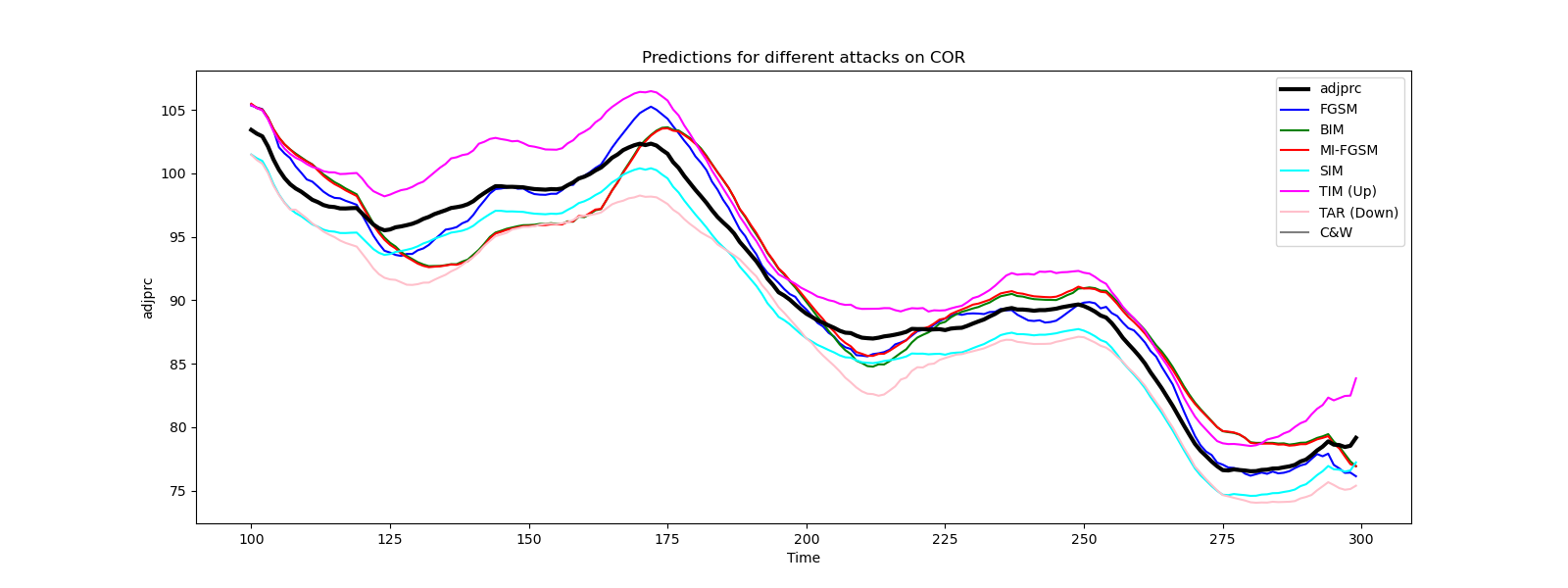}
    \caption{Various attacks performed on the stock \textit{COR}. Attacks were performed on the first 300 days of the recording.}   
    \label{fig:cor_attack_example}
\end{figure}
\FloatBarrier

\subsection{C\&W Slope Attack Example}
\begin{figure}[H]
    \centering
    \includegraphics[scale=0.35]{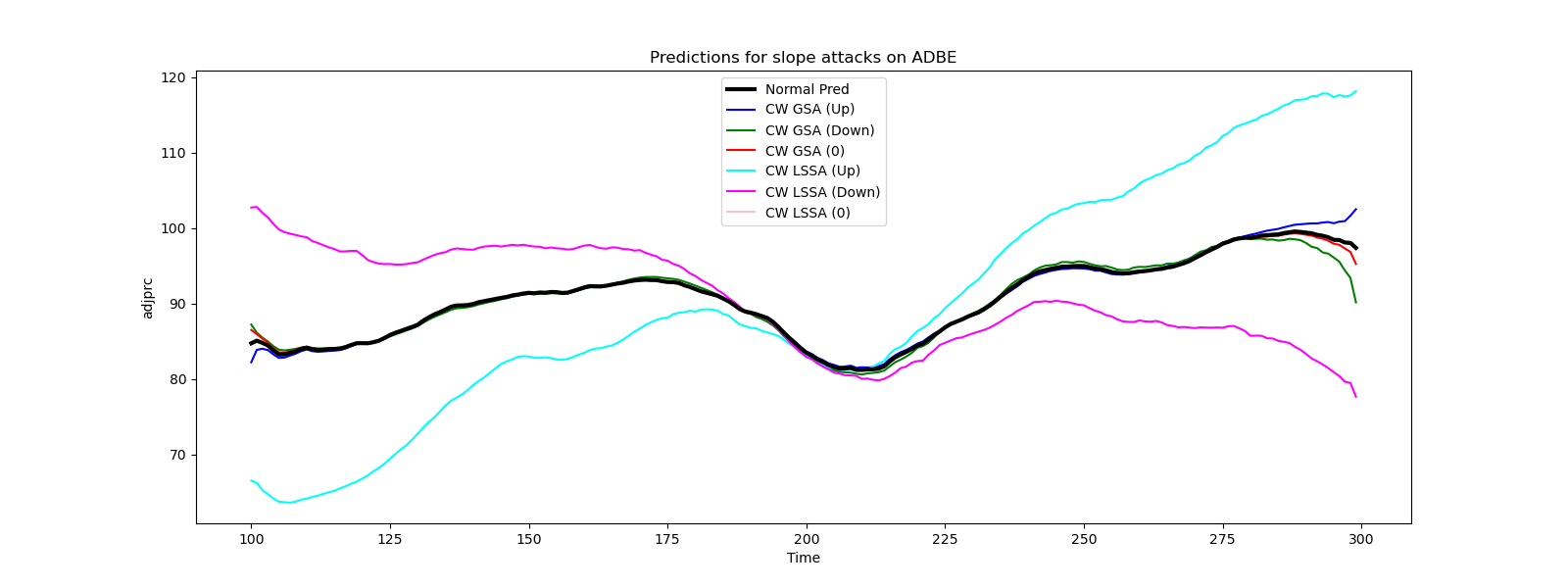}
    \caption{C\&W slope attacks performed on the stock \textit{ABDE}. Attacks were performed on the first 300 days of the recording.}   
    \label{fig:cw_slope_attack}
\end{figure}
\FloatBarrier

\subsection{Baseline GAN Explanations}

\subsubsection{DCGAN}
The DCGAN is an extension of the original GAN, where the generator and discriminator are made using transpose and normal convolutions \cite{dcgan:2015}. The generator takes a small noise vector as input, and passes the vector through multiple strided transpose convolutions in order to increase the dimensionality to match the real data \cite{dcgan:2015}. The discriminator typically performs the opposite transformation, using strided convolutions to reduce the dimensionality to a single value, representing the probability of the sample being real or fake \cite{dcgan:2015}.

\subsubsection{WGAN}
In practice, vanilla GANs are typically notoriously hard to train, due to the possible discontinuities between the divergence the generator optimizes and the generators parameters \cite{wgangp:2017}. Thus, the objective function was changed to the approximate the Wasserstein distance which, under mild assumptions, is continuous everywhere \cite{wgangp:2017}. Specifically, the objective function is defined as:
\begin{align*}
    \min_G \max_D \mathbb{E}_{x\sim p_{data}(x)}[D(x)] - \mathbb{E}_{z \sim p_{z}(z)}[D(G(z))] \text{\quad \quad \cite{wgangp:2017}}
\end{align*}
However, the original implementation of the WGAN enforced the Lipchitz constraint using gradient clipping, which can be problematic due to vanishing gradients and optimization difficulties \cite{wgangp:2017}. Therefore, a gradient penalty term was added to the objective function to overcome the optimization difficulties, defined as 
\begin{align*}
    L = \mathbb{E}_{\tilde{x} \sim P_g}[D(\tilde{x})] - \mathbb{E}_{x \sim P_r}[D(x)] + \lambda \cdot \mathbb{E}_{\hat{x} \sim P_{\hat{x}}} [(||\nabla_{\hat{x}} D(\hat{x})|| - 1)^2]
\end{align*}
where $\tilde{x}$ is a synthetic sample, and $\lambda$ controls the magnitude of the gradient penalty \cite{wgangp:2017}.

\subsection{Sample A-GAN Outputs}
\begin{figure}[H]
    \centering
    \includegraphics[scale=0.6]{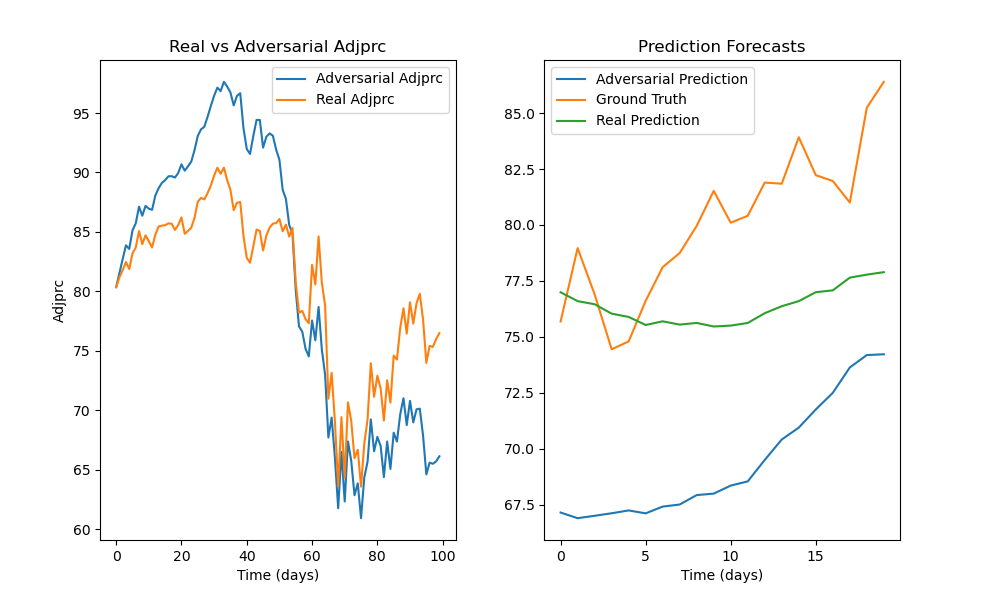}
    \vspace{-2mm}
    \caption{Example A-GAN output generated from a random interval.}
    \label{fig:a_gan_output_12}
\end{figure}
\begin{figure}[H]
    \centering
    \includegraphics[scale=0.6]{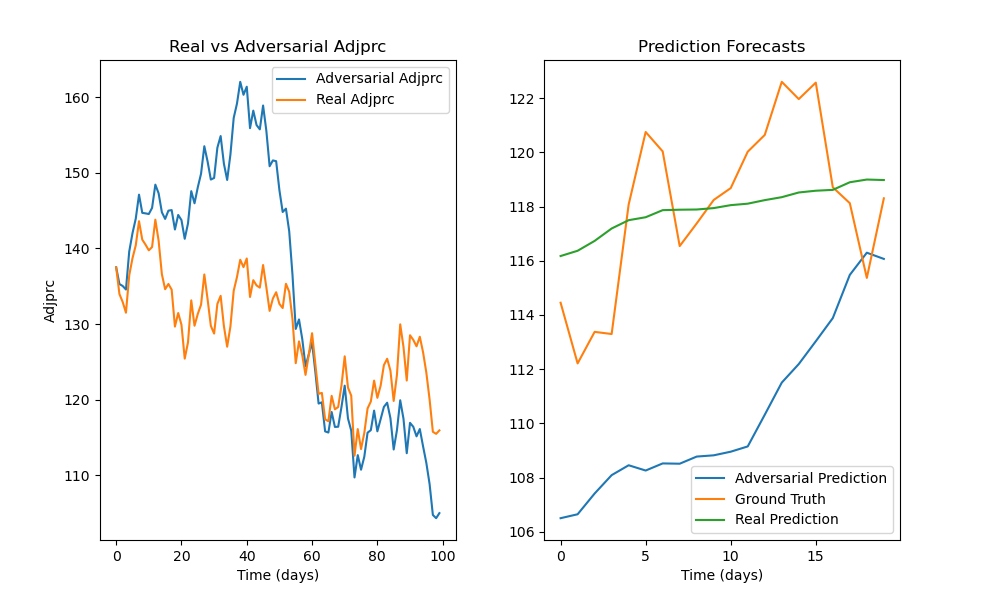}
    \vspace{-2mm}
    \caption{Example A-GAN output generated from a random interval.}
    \label{fig:a_gan_output_15}
\end{figure}
\begin{figure}[H]
    \centering
    \includegraphics[scale=0.6]{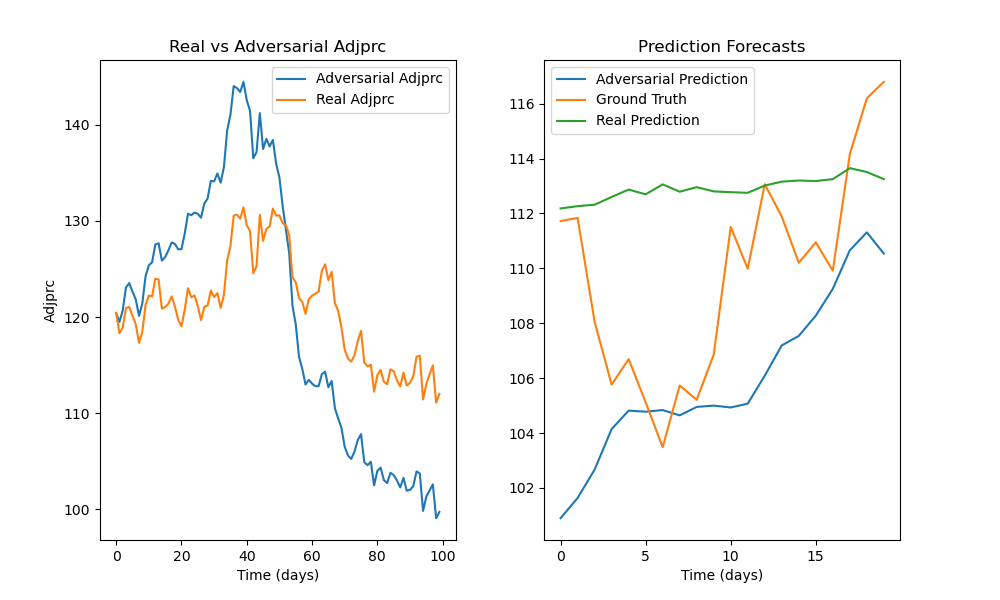}
    \vspace{-2mm}
    \caption{Example A-GAN output generated from a random interval.}
    \label{fig:a_gan_output_18}
\end{figure}
\FloatBarrier
\subsection{Model Hyperparameters}
\begin{table}[H]
    \centering
    \caption{Hyperparameters used for the N-HiTS model}
    \vspace{-3mm}
    \begin{tabular}{cc}
        \hline
        \textit{Hyperparameter} & \textit{Value}\\
        \hline
        \text{Learning Rate} & $10^{-3}$ \\
        \text{Weight Decay} & $10^{-4}$ \\
        \text{Hidden Size} & 64 \\
        \text{Batch Normalization} & True \\
        \text{Early Stopping} & True \\
        \text{Max/Min Encoder Length} & 100\\
        \text{Max/Min Prediction Length} & 20\\
        \text{Batch Size} & 500 \\
        \text{Epochs} & 100\\
        \hline
    \end{tabular}
    \label{tab:N-HiTS_hparams}
\end{table}
\FloatBarrier
\begin{table}[H]
    \centering
    \caption{Hyperparameters used for the CNN Discriminator}
    \vspace{-3mm}
    \begin{tabular}{cc}
        \hline
        \textit{Hyperparameter} & \textit{Value}\\
        \hline
        \text{Learning Rate} & $10^{-4}$ \\
        \text{Weight Decay} & $10^{-5}$ \\
        \text{Hidden Size} & (16, 32, 16) \\
        \text{Batch Normalization} & True \\
        \text{Batch Size} & 32 \\
        \text{Dropout} & 0.2 \\
        \text{Epochs} & 200\\
        \hline
    \end{tabular}
    \label{tab:discrim_h_params}
\end{table}
\FloatBarrier
\begin{table}[H]
    \centering
    \caption{Hyperparameters used for the A-GAN Model}
    \begin{tabular}{cc}
        \hline
        \textit{Hyperparameter} & \textit{Value}\\
        \hline
        \text{Batch Size} & 32 \\
        \text{Samples Per Epoch} & 512\\
        \text{Number of Critic Iterations} & 5 \\
         $\lambda$ & 1 \\
        \text{Generator Learning Rate} & $10^{-4}$ \\
        \text{Critic Learning Rate} & $10^{-4}$ \\
        \text{Betas} & (0, 0.9) \\
        \text{Dropout} & 0.2 \\
        \text{Batch Normalization} & False \\
        \text{Leaky ReLU Negative Slope} & 0.2\\
        \text{Generator Number of TCN Layers} & 4\\
        \text{Generator Hidden Size} & (64, 128, 64, 32)\\
        \text{Generator TCN Kernel Sizes} & (3, 5, 5, 3)\\
        \text{Generator TCN Dilation} & (1, 2, 4, 8)\\
        \text{Critic Number of Layers} & 5\\
        \text{Critic Number of TCN Layers} & 3\\
        \text{Critic Number of GRU Layers} & 2\\
        \text{Critic Hidden Size} & (64, 128, 128, 64, 32)\\
        \text{Critic TCN Kernel Sizes} & (3, 5, 5)\\
        \text{Critic TCN Dilation} & (1, 2, 4)\\
        \text{Epochs} & (50, 50, 50, 50, 50)\\
        \text{$\alpha$} & (0.25, 0.25, 0.3, 0.35, 0.35)\\
        \text{c} & 5 \\
        \text{d} & 2 \\
        \hline
    \end{tabular}
    \label{tab:wgan_h_params}
\end{table}
\FloatBarrier

\clearpage
\printbibliography

\end{document}